\documentclass[conference]{IEEEtran}
\IEEEoverridecommandlockouts
\usepackage{cite}
\usepackage{amsmath,amssymb,amsfonts}
\usepackage{algorithmic}
\usepackage[ruled,vlined,linesnumbered]{algorithm2e}
\usepackage{graphicx}
\usepackage{textcomp}
\usepackage{xcolor}
\usepackage{multirow}
\usepackage{multicol}
\def\BibTeX{{\rm B\kern-.05em{\sc i\kern-.025em b}\kern-.08em
    T\kern-.1667em\lower.7ex\hbox{E}\kern-.125emX}}

\newtheorem{example}{Example}
\newtheorem{problem}{Problem}

\usepackage{xcolor}

\usepackage{paralist}
\usepackage{url}

\begin{document}

\title{DDTR: Diffusion Denoising Trace Recovery}

\author{\IEEEauthorblockN{Maximilian Matyash}
\IEEEauthorblockA{\textit{Data \& Decision Sciences} \\
\textit{Technion-Israel Institute of Technology}\\
Haifa 3200003, Israel \\
maximilian.m@campus.technion.ac.il \\
orcidID: 0009-0008-8304-1188
}
\and
\IEEEauthorblockN{Avigdor Gal}
\IEEEauthorblockA{\textit{Data \& Decision Sciences} \\
\textit{Technion-Israel Institute of Technology}\\
Haifa 3200003, Israel \\
avigal@technion.ac.il \\
orcidID: 0000-0002-7028-661X
}
\and
\IEEEauthorblockN{Arik Senderovich}
\IEEEauthorblockA{\textit{School of Information Technology}} 
\textit{York University}\\
Toronto, Canada \\
sariks@yorku.ca \\
orcidID: 0000-0003-4728-8024
}

\maketitle

\begin{abstract}
With recent technological advances, process logs, which were traditionally 
deterministic in nature, are being captured from non-deterministic sources, 
such as uncertain sensors or machine learning models (that predict activities 
using cameras). 
In the presence of stochastically-known logs,
logs that contain probabilistic information, 
the need for stochastic trace recovery increases, to offer reliable means of understanding the processes 
that govern such systems. We design a novel deep learning approach for stochastic trace recovery, based on Diffusion Denoising Probabilistic Models (DDPM), which makes use of process knowledge (either implicitly by discovering a model or explicitly by injecting process knowledge in the training phase) to recover traces by denoising. 
We conduct an empirical evaluation demonstrating state-of-the-art performance with up to a 25\% improvement over existing methods, along with increased robustness under high noise levels. 
\end{abstract}

\section{Introduction}
The goal of process mining is to discover, analyze, and optimize real-world processes~\cite{DBLP:books/sp/Aalst16}. 
Servicing a patient in a hospital, filing a marriage certificate, or 
preparing a meal are all examples of real-world processes that can be analyzed and improved.
Process mining relies on a process log, a recording of process execution,
which contains a collection of traces where each trace is a sequence of activities. 
Traditionally, a process log is created by either manually logging real-world activities ({\em e.g.}, a nurse keying in the timestamp of finishing an examination), or having activities automatically captured and logged by an information system.  

Process logs are assumed to record events in a deterministic fashion. However, modern means of recording processes have emerged, which  result in stochastically-known process logs~\cite{everything_sklogs}. Consider the case of a smart camera, which we shall use as a running example. The camera records footage of a room and uses a machine learning (ML) classification model to determine filmed activities. The resulting process log is comprised of stochastic traces, each a sequence of probability distributions over all possible activities, deduced from the final softmax layer of the classification model. Transforming the stochastic log into a deterministic one enables the use of existing process mining tools. Any such transformation must aim to represent what truly happened in the recorded process.

The task of \textit{trace recovery} was introduced by Bogdanov {\em et al.}~\cite{sktr}, showing that the na\"ive approach of selecting
the activity with the highest probability at each time step (known as the argmax approach) is prone to misrepresenting the log in reality. In previous work \cite{sktr}, this problem was solved by examining the stochastic log in the context of the system's process model and providing an alignment-based solution for trace recovery that outperforms the na\"ive approach. 

In this work, we reframe the trace recovery problem as a machine learning task and introduce a novel approach based on Diffusion Denoising Probabilistic Models (DDPM)~\cite{ddpm}. Originally developed for self-supervised image generation, DDPMs learn data distributions over structured spaces, making them surprisingly suitable for modeling event logs. We show how DDPMs can recover deterministically known traces from stochastically observed logs with minimal supervision.

We present Diffusion Denoised Trace Recovery (DDTR), which extends the idea of \emph{classifier-free guidance}~\cite{classifierfree}, where generation is influenced by auxiliary signals. We analogize trace recovery to image deblurring: a noisy or partial trace is iteratively refined into a deterministic one under the guidance of its probabilistic origin. We further demonstrate how process models can serve as additional guidance, improving recovery performance.
The paper makes the following contributions:
\begin{compactitem}
    \item We position trace recovery from stochastically-known logs as a generative denoising problem, uncovering a novel link between log and image reconstruction.
    \item We build upon existing DDPM methods to solve both model-aware and model-free stochastic trace recovery problems, the latter of which we define in this work. 
    \item We provide a comprehensive empirical analysis of our model's performance at various levels of noise of the stochastic traces and demonstrate that incorporating process models into the method achieves high robustness in cases where noise levels are high. In addition, we show that our method achieves a relative accuracy improvement of 5\%-25\% over state-of-the-art methods on well-established real-world benchmark datasets.
\end{compactitem} 

The rest of the paper is organized as follows. We introduce diffusion denoising models in Section~\ref{sec:Background}, give an overview for our approach and notation (Section~\ref{sec:Model}), present DDTR (Section~\ref{sec:method}) and discuss empirical results (Section~\ref{sec:eval}). We discuss relevant related work in Section~\ref{sec:rel} before concluding the paper in Section~\ref{sec:conclusion}.

\section{Diffusion Denoising Probabilistic Models}
\label{sec:Background}
Diffusion denoising probabilistic models (DDPM) are self-supervised generative models, originally developed to improve image generation~\cite{diffusionbeatgan} over generative adversarial networks~\cite{gan} (GAN) and variational auto-encoders~\cite{vae} (VAE). 
Given a set of datapoints $\mathcal{X}=\{\vec{x_0}^{(i)}|i=1,\dots,M\}$ (for example, images), these generative models sample new examples from the distribution $p(\vec{x}_0)$ using two components, namely a {\em forward process} (Section~\ref{sec:forward}) and a {\em reverse process} (Section~\ref{sec:reverse}). Table~\ref{tab:notation} summarizes the notation we use in this work. 

\subsection{Forward Process}\label{sec:forward}
The forward process gradually adds noise to a datapoint $\vec{x_0}$ by mixing it with Gaussian noise as follows.
\begin{equation}
\label{eq:forward_proc}
    \vec{x}_t = \sqrt{\alpha_t}\vec{x}_{t-1}+\sqrt{1-\alpha_t}\varepsilon;~\varepsilon\sim\mathcal{N}(\vec{0},\mathbf{I})
\end{equation}
where $\{\alpha_t\}_1^T$ are hyperparameters called the \textit{noise schedule} and $\vec{x}_t$ is a perturbation of $\vec{x}_0$ at time $t$, meaning that the original datapoint was mixed with noise $t$ times to create $\vec{x}_t$. The process is applied for $T$ steps per datapoint, creating a sequence of random variables $\vec{x}_1,\dots,\vec{x}_T$ whose probability distributions are defined as $q(\vec{x}_t|\vec{x}_{t-1}):=\mathcal{N}(\sqrt{\alpha_t}\vec{x}_{t-1},(1-\alpha_t)\mathbf{I})$. The final variable $\vec{x}_T$ becomes pure Gaussian noise ($\vec{x}_T\sim\mathcal{N}(\vec{0},\mathbf{I})$), holding no information about the original datapoint. Therefore, the forward process is destructive, creating continuous paths from an unknown distribution $p(\vec{x}_0)$ to a known distribution $p(\vec{x}_T)=\mathcal{N}(\vec{0},\mathbf{I})$. 

\subsection{Reverse Process}\label{sec:reverse}
The reverse process starts with random Gaussian noise $\vec{x}_T\sim\mathcal{N}(\vec{0},\mathbf{I})$ and iteratively generates a reverse path $\hat{x}_{T-1},\dots,\hat{x}_0$ which ends with a sample from the approximated distribution of $\mathcal{X}$, $\hat{p}(\vec{x}_0)$. This is achieved by training a {\em denoiser}, a deep neural network $\hat{x}_0(x_t,t;\theta)$, where $\theta$ are the network's parameters, as described in Eq.~\ref{eq:back_proc}.
\begin{equation}
\label{eq:back_proc}
\resizebox{1.0\linewidth}{!}{%
$\begin{cases}
    \hat{x}_{t-1} = 
      \frac{\sqrt{\bar{\alpha}_{t-1}}}{1-\bar{\alpha}_t}
         \hat{x}_0(\hat{x}_t,t;\theta)+
      \frac{\sqrt{\alpha_t}(1-\bar{\alpha}_{t-1})}{1-\bar{\alpha}_t}
         \hat{x}_t+
      \frac{1-\bar{\alpha}_{t-1}}{1-\bar{\alpha}_t}(1-\alpha_t)z
      \\[6pt]
    z \sim \mathcal{N}(\vec{0},\mathbf{I})
      ~\text{if}~t>0
      ~\text{else}~z=\vec{0}
\end{cases}$}%
\end{equation}
$\bar{\alpha}_t:=\prod_{s=1}^t\alpha_s$. The denoiser predicts the original datapoint for a given noisy instance.

\begin{table}[t]
    \centering
        \caption{Summary of Notation}
    \begin{tabular}{c|c}
         \hline
         Symbol & Meaning \\
         \hline
         $\mathbf{D^{(i)}}$ & Deterministically known (DK) trace matrix representation \\
         $\mathbf{S^{(i)}}$ & Stochastically known (SK) trace matrix representation \\
         $T_i$ & Length of trace $i$ \\
         $\tau$ & Timestep within trace \\
         $t$ & Timestep within diffusion process \\
         $T$ & Number of diffusion steps \\
         $x_0$ & Source datapoint, DK trace \\
         $x_t$ & Noised datapoint within the forward process \\
         $y, z$ & Guiding data - SK trace, process model structure \\
         $\hat{x}_0(x_t,t,y,z;\theta)$ & Denoiser output - predicted DK trace \\
         $\hat{x}_{t}$ & Denoised datapoint within the reverse process \\
         $\theta$ & Denoiser's parameters \\
         $\alpha_t$ & Diffusion process noise schedule constants \\
         $\bar{\alpha}_t$ & Diffusion process cumulative noise constants \\
    \end{tabular}
    \label{tab:notation}
\end{table}

The reverse process aims to invert the perturbations of the forward process and therefore needs to be stochastic in nature as well, which is why Gaussian noise is still injected during the reverse process through the random variable $z$, except for the final step which yields the sampled datapoint.

Typically, the denoiser in DDPM predicts the noise in a given instance instead of the original datapoint. However, Dieleman {\em et al.}~\cite{diffusionce} show that in highly discrete cases (such as processes with a limited number of activities) predicting the original datapoint results in better performance. For this reason we reformulate the reverse process equation in terms of $x_0$ via simple variable substitution.

\begin{figure}[htpb]
    \centering
    \includegraphics[width=0.8\linewidth]{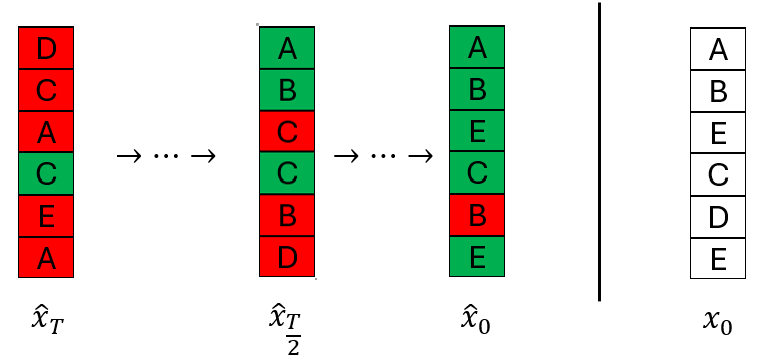}
    \caption{Illustration of the trace recovery process in DDTR. The recovery process begins on the left, with a random guess, denoted $\hat{x}_T$. The guess then goes through an iterative process of refinement and results in a highly accurate recovered trace $\hat{x}_0$. The trace denoted $x_0$ is the ground truth DK trace.}
    \label{fig:denoising_example}
\end{figure}

Figure~\ref{fig:denoising_example} shows an illustration of the reverse process applied to a single trace. The process starts on the left with $\hat{x}_T$ which is a random sequence of activities. The denoiser is invoked $T$ times using Eq.~\ref{eq:back_proc}, resulting in an approximated process trace $\hat{x}_0$.

The denoiser is trained over multiple epochs by randomly selecting a timestep $t$ and having the denoiser predict the original input given $x_t$ and then evaluating the prediction with a loss function $\mathcal{L}(x_0,\hat{x}_0(x_t,t;\theta))$. The denoiser's parameters are updated via the gradients of the loss function through backpropagation and the process is repeated for a large number of epochs or until convergence.

\begin{figure}[htpb]
    \centering
    \includegraphics[width=1\linewidth]{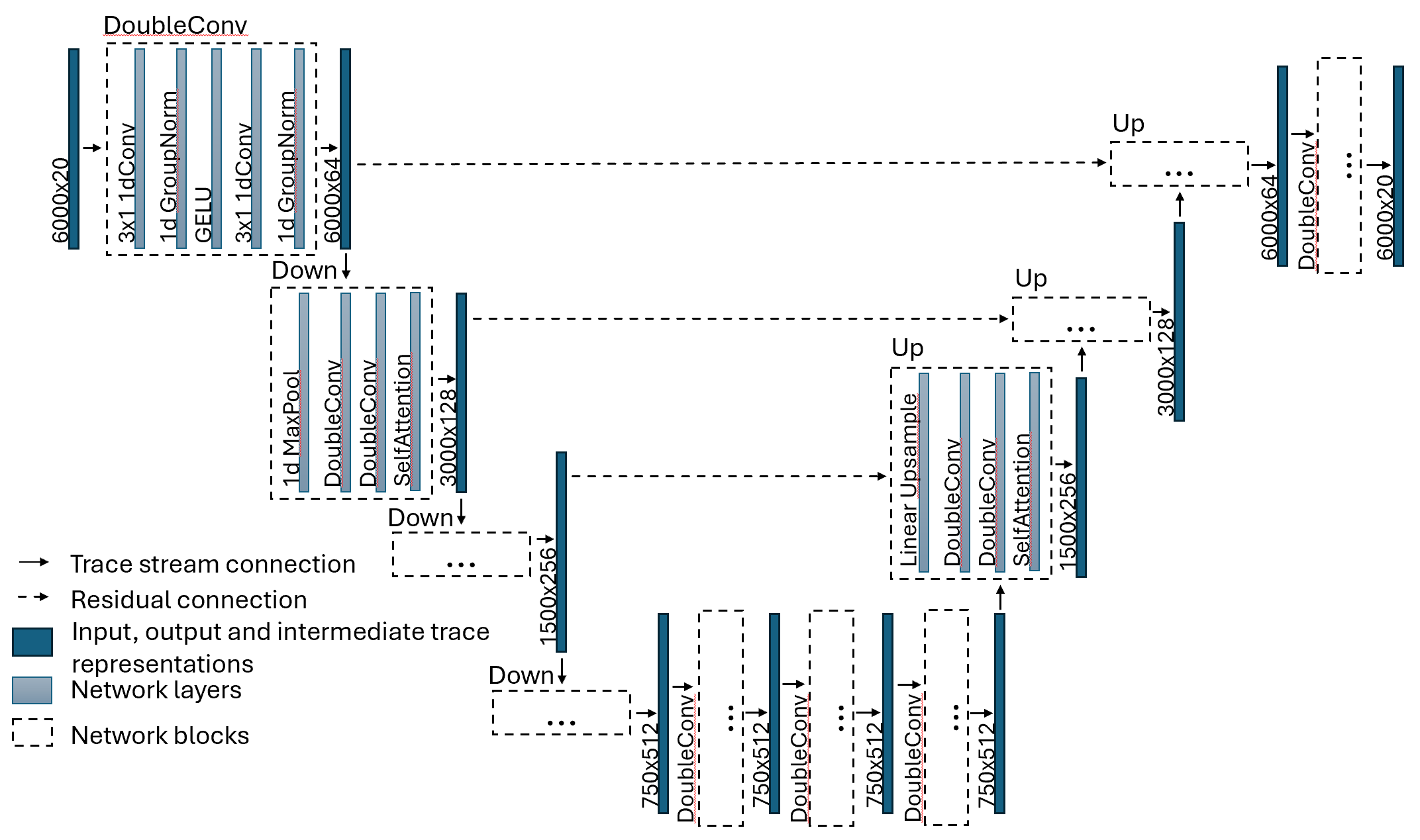}
    \caption{Architecture of a U-net denoiser for trace generation. From left to right: the input trace is the result of the previous step in the reverse process, it goes through a series of convolutions and attention layers which generate features that are used to reconstruct a refined prediction of the DK trace.}
    \label{fig:unet_base}
\end{figure}

The commonly used architecture for the denoiser neural network is in the form of a U-net~\cite{unet}, which is a convolutional neural network that is made of two symmetric paths of downsampling and upsampling. The architecture of a base U-net is illustrated in Figure~\ref{fig:unet_base}. Starting from the top left, the downward path consists of convolutions that increase the number of features of decreased size, achieved by max pooling. On the right, the upward path consists of de-convolutions that increase in size symmetrically to the downward path, matching the output and input shape. At the bottom, the two paths are connected by a string of convolutions, called the {\em bottleneck}, that learns the most high-level features and bridges the two paths. In addition, skip connections between layers on the downward path and their counterparts on the upward path provide the upsampling additional context that might be lost in the downsampling path.

\subsection{Guidance and Inverse Problems}
\label{subsec:guide}
Using DDPM on its own allows for unconditional generation of new datapoints by sampling from the learned distribution $\hat{p}(x_0)$. However, having control over the resulting sample, by conditioning on some external information, may be desirable. For example, in the task of trace recovery we do not wish to just sample any deterministically known trace but rather the specific recovered version of an input stochastically known trace.

\textit{Guidance}\cite{classifierfree} supports conditional generation using DDPM. Given an external variable $y$, the denoiser is modified to take $y$ as input, $\hat{x}_0(x_t,t,y;\theta)$, allowing sampling from the learned conditional distribution of datapoints $\hat{p}(x_0|y)$.

Beyond conditional generation, guided diffusion models have been used to solve inverse problems, primarily in the domain of images. \textit{Inverse problems} are formulated by the equation $\vec{y}=f(\vec{x})$, with the goal of extracting $\vec{x}$ given $\vec{y}$. The transformation $f(\vec{x})$ is mostly unknown or difficult to model and guided diffusion models assist by bypassing the transformation to directly learn and then sample solutions from the posterior probability $\hat{p}(\vec{x}|\vec{y})$. Examples for such works can be seen in deblurring, super-resolution~\cite{ddrm}, phase retrieval~\cite{diffusion_general_inverse}, inverse rendering~\cite{diffusion_inverse_render}, and optical flow estimation~\cite{diffusion_flow_estimation}. 

The main challenge in successfully training a guided diffusion model lies in the architecture of the denoiser neural network, especially in finding the correct way to combine $x_t$ and $y$ across the various layers of the network.


\section{Data Model and Problem Definition}
\label{sec:Model}

Stochastically known (SK) traces \cite{uncertainlogs}\cite{everything_sklogs} describe uncertainty of an underlying deterministic process. The primary characterization of SK traces is that each event is a probability distribution over all possible activities. Deterministically known (DK) traces, the typical traces encountered in process mining, serve as a special case of SK where each event is deterministically mapped to an activity. That is, a DK trace is a SK trace where a single activity has a probability $1$ of being reported, while other activities have a probability of $0$.

\begin{table}[htpb]
    \centering
    \caption{An excerpt from a DK log. Each event is mapped to a single activity from the finite set of all activities}
    \begin{tabular}{cccc}
         \hline
         Case Id & Event no. & Activity & Relative Time \\
         \hline
         1 & $e_1$ & A & 00:00 \\
         1 & $e_2$ & B & 00:10 \\
         1 & $e_3$ & E & 00:40 \\
         1 & $e_4$ & C & 01:10 \\
         1 & $e_5$ & D & 01:30 \\
         1 & $e_6$ & E & 01:45 \\
    \end{tabular}
    
    \label{tab:dk_trace}
\end{table}
\begin{table}[htpb]
    \centering
    \caption{An excerpt from a SK log. Each event is mapped to a probability vector over the finite set of all activities}
    \begin{tabular}{cccc}
         \hline
         Case Id & Event no. & Activity & Relative Time \\
         \hline
         1 & $e_1$ & (0.33, 0.03, 0.15, 0.15, 0.34) & 00:00 \\
         1 & $e_2$ & (0.2, 0.25, 0.15, 0.2, 0.2) & 00:10 \\
         1 & $e_3$ & (0.5, 0.1, 0.1, 0.1, 0.2) & 00:40 \\
         1 & $e_4$ & (0.05, 0.15, 0.55, 0.05, 0.2) & 01:10 \\
         1 & $e_5$ & (0.1, 0.05, 0.25, 0.45, 0.15) & 01:30 \\
         1 & $e_5$ & (0.1, 0.05, 0.25, 0.25, 0.35) & 01:45 \\
    \end{tabular}
   
    \label{tab:sk_trace}
\end{table}

\begin{example}
Consider the trace marked with $x_0$ in Figure~\ref{fig:denoising_example}. Table~\ref{tab:dk_trace} illustrates an excerpt from a DK log of this trace and Table~\ref{tab:sk_trace} illustrates an excerpt of the same trace from a SK log for a process with five activities, {\em A, B, C, D} and {\em E}. 
\end{example}

The assumption in our setting is that the process in reality is deterministic; 
however, we have access only to a limited number 
of DK traces and otherwise we can only observe SK traces. For each 
observed SK trace there exists a ground truth DK trace that we aim to recover. 

A na\"ive approach for trace recovery would be to independently take the activity with the highest probability for each event (we will refer to this as the \textit{argmax} approach). Yet, by doing so, the inherent biases and 
inaccuracies of the source of the SK logs are preserved in our DK reconstruction. For example, when taking the argmax in Table~\ref{tab:sk_trace}, we get the activities $({\bf E},B,{\bf A},C,D,E)$ while the true activities as in Table~\ref{tab:dk_trace} are $({\bf A},B,{\bf E},C,D,E)$.

This approach can be improved by taking into account the context of the entire process model, which provides key information about activity viability at each time point. This was shown 
to yield improved results over the argmax approach~\cite{sktr}.
In this work, we use deep learning to improve the predictive power of the recovery process.
We position the task of stochastic trace recovery as an inverse problem (Section~\ref{subsec:guide}) and solve it with a guided diffusion model. 

Let $D^{(i)}=\langle D_1^{(i)},\dots,D_{T_i}^{(i)}\rangle$ be a DK trace of length $T_i$ from a DK process log. Let $\mathcal{A}=\{a_1,...,a_K\}$ be the set of all possible activities in the process, ordered arbitrarily from activity 1 to activity $K$. Events in $D^{(i)}$ correspond to activities in $\mathcal{A}$, meaning that $\forall\tau\in\{1,\dots,T_i\}:D_{\tau}^{(i)}\in\mathcal{A}$. We represent each activity with a unique deterministic probability vector in the set $\mathcal{A}_{\delta}=\{\vec{a_i}\in\{0,1\}^K  \ | \ \sum_{j=1}^K \vec{a_i}^{(j)}=1 \}$ such that the entire DK trace can be represented as the matrix $\mathbf{D^{(i)}}$, which columns are the probability vectors corresponding to the activities in $D^{(i)}$. 

Let $S^{(i)}=\langle S_1^{(i)},\dots,S_{T_i}^{(i)}\rangle$ be a SK trace and $\mathcal{S}=\{s\in\mathbb{R}^K \ | \ \sum_{j=1}^Ks_i=1\}$ the set of probability vectors over activities such that $\forall\tau\in\{1,\dots,T_i\}:S_{\tau}^{(i)}\in\mathcal{S}$ and, similarly to the DK trace, the SK trace can be represented as a matrix $\mathbf{S^{(i)}}$ whose columns are the respective probability vectors.

\begin{example}
The matrices below are the respective representations for the DK and SK trace excerpts of tables \ref{tab:dk_trace} and \ref{tab:sk_trace}: \\
\[
\mathbf{D^{(i)}}=~
\begin{array}{c}
    A \\
    B \\
    C \\
    D \\
    E
\end{array}
\begin{bmatrix}
    1 & 0 & 0 & 0 & 0 & 0 \\
    0 & 1 & 0 & 0 & 0 & 0 \\
    0 & 0 & 0 & 1 & 0 & 0 \\
    0 & 0 & 0 & 0 & 1 & 0 \\
    0 & 0 & 1 & 0 & 0 & 1
\end{bmatrix}
\]
\[
\mathbf{S^{(i)}}=~
\begin{array}{c}
    A \\
    B \\
    C \\
    D \\
    E
\end{array}
\begin{bmatrix}
    0.33 & 0.2 & 0.5 & 0.05 & 0.1 & 0.1 \\
    0.03 & 0.25 & 0.1 & 0.15 & 0.05 & 0.05 \\
    0.15 & 0.15 & 0.1 & 0.55 & 0.25 & 0.25 \\
    0.15 & 0.2 & 0.1 & 0.05 & 0.45 & 0.25 \\
    0.34 & 0.2 & 0.2 & 0.2 & 0.15 & 0.35
\end{bmatrix}
\]
\end{example}

The data set can be represented as $\mathcal{D}=\{(\mathbf{D^{(i)}},\mathbf{S^{(i)}})\ \vert \ i=1,\dots,N\}$, consists of $N$ pairs of a SK trace and its ground truth DK trace. Ground truth is known to the model only during training. During inference the model receives a SK trace as input and predicts the recovered DK trace.

We model the problem with a Dirichlet mixture~\cite{dirichlet}, which mixes the DK trace with stochastic transition vectors that represent random transition probabilities to all activities, regardless of whether they are permitted in the process.

Consider that for a single deterministic activity at time $\tau$, we can transition to a stochastic activity by mixing it with random Dirichlet noise $\vec{\pi}\sim\text{Dirichlet}(\vec{\alpha})$ at some noise level $\lambda$: $\vec{R}_{\tau}^{(i)}=(1-\lambda)\vec{D}_{\tau}^{(i)}+\lambda_{\tau}\vec{\pi}$ and re-normalizing: 
\begin{equation}
\vec{S}_{\tau}^{(i)}=\frac{\vec{R}_{\tau}^{(i)}}{\sum_{j=1}^K \vec{R}_{\tau,j}^{(i)}}.
\end{equation}
We assume that the noise level can differ over the course of the trace's execution, thus the mixture parameter $\lambda$ is dependent on the trace's time step $\tau$. Extending this notion to all of the activities comprising the DK trace can be written as $\mathbf{D}^{(i)}=\langle \vec{D}_1^{(i)},\dots,\vec{D}_{\tau}^{(i)},\dots,\vec{D}_{T_i}^{(i)}\rangle$,
and we can write the generation process of SK trace as follows. \begin{equation}
\label{eq:det_to_stoch}
    \begin{cases}
        \mathbf{R^{(i)}}=(\mathbf{I}-\mathbf{\Lambda})\mathbf{D^{(i)}}+\mathbf{\Lambda}\mathbf{\Pi},\\
        \mathbf{S^{(i)}}=\mathbf{R^{(i)}}\text{diag}(\vec{1}^T\mathbf{R^{(i)}})^{-1},
    \end{cases}
\end{equation} where $\mathbf{\Lambda}:=\text{diag}(\lambda_1,\dots,\lambda_{T_i})$, $\mathbf{\Pi}:=\langle\vec{\pi}_1,\dots,\vec{\pi}_{T_i}\rangle$ is a matrix created by concatenating independent random Dirichlet variables $\vec{\pi}_1,\dots,\vec{\pi}_{T_i}$ as columns and $\mathbf{I}$ is the identity matrix, the remaining operations on $\mathbf{R}^{(i)}$ divide each column by its sum, thus resulting in legal probability distributions. $\vec{1}$ is the unit vector of 
the appropriate length.
The resulting formulation is a non-linear set of equations since normalization is a non-linear operation. 

We are now ready to define the trace recovery problems. 
\begin{problem}[Model-Free Trace Recovery] \label{prob1}
The model-free trace recovery problem is to find the true $\mathbf{D}^{(i)}, \forall i \in \{1, \ldots, N\}$,
based on the SK $\mathbf{S^{(i)}}$.
\end{problem} 


In addition to the stochastic trace we can make use of 
the structure of a previously (automatically or manually) discovered Petri net (or any process model with a flow matrix) as an additional source of guidance. For example, a model can be a Petri net $N=(P,T,F)$ mined from the DK log, with $F$ being the flow relation matrix which holds all edges corresponding to the directed graph induced by $N$. We can extend Problem~\ref{prob1} to a
model-based trace recovery problem, which was introduced by Bogdanov {\em et al.}~\cite{sktr}. 
\begin{problem}[Model-Based Trace Recovery] \label{prob2}
The model-based trace recovery problem is to find the true $\mathbf{D}^{(i)}, \forall i \in \{1, \ldots, N\}$,
based on the SK $\mathbf{S^{(i)}}$ and a flow matrix of some process model $F$.
\end{problem} 

The diffusion process 
that we use to solve these problems is extended with process structure information that allows sampling from $\hat{p}(\mathbf{D}|\mathbf{S},\tilde{F})$, where $\tilde{F}$ is a latent representation of $F$, learned by the denoiser. In case we do not have a model, we write $\hat{p}(\mathbf{D}|\mathbf{S})$ instead.

It is worth noting that Problem~\ref{prob1} is a generalization of Problem~\ref{prob2} and is inherently more difficult to solve since the model, which holds crucial information, is absent.

\section{Method}
\label{sec:method}
DDTR is a trace recovery method, implemented as a guided diffusion process. We designed a custom implementation of a trace denoiser that uses process model structure as input, returning a denoised trace and a process' flow matrix. In Section~\ref{sec:network}, we demonstrate the gradual creation of the denoiser neural network structure. Section~\ref{sec:algs} presents algorithms for training a denoiser and generating recovered traces.


\subsection{Denoiser Neural Network}
\label{sec:network}
\begin{figure}[t]
    \centering
\includegraphics[width=0.9\linewidth]{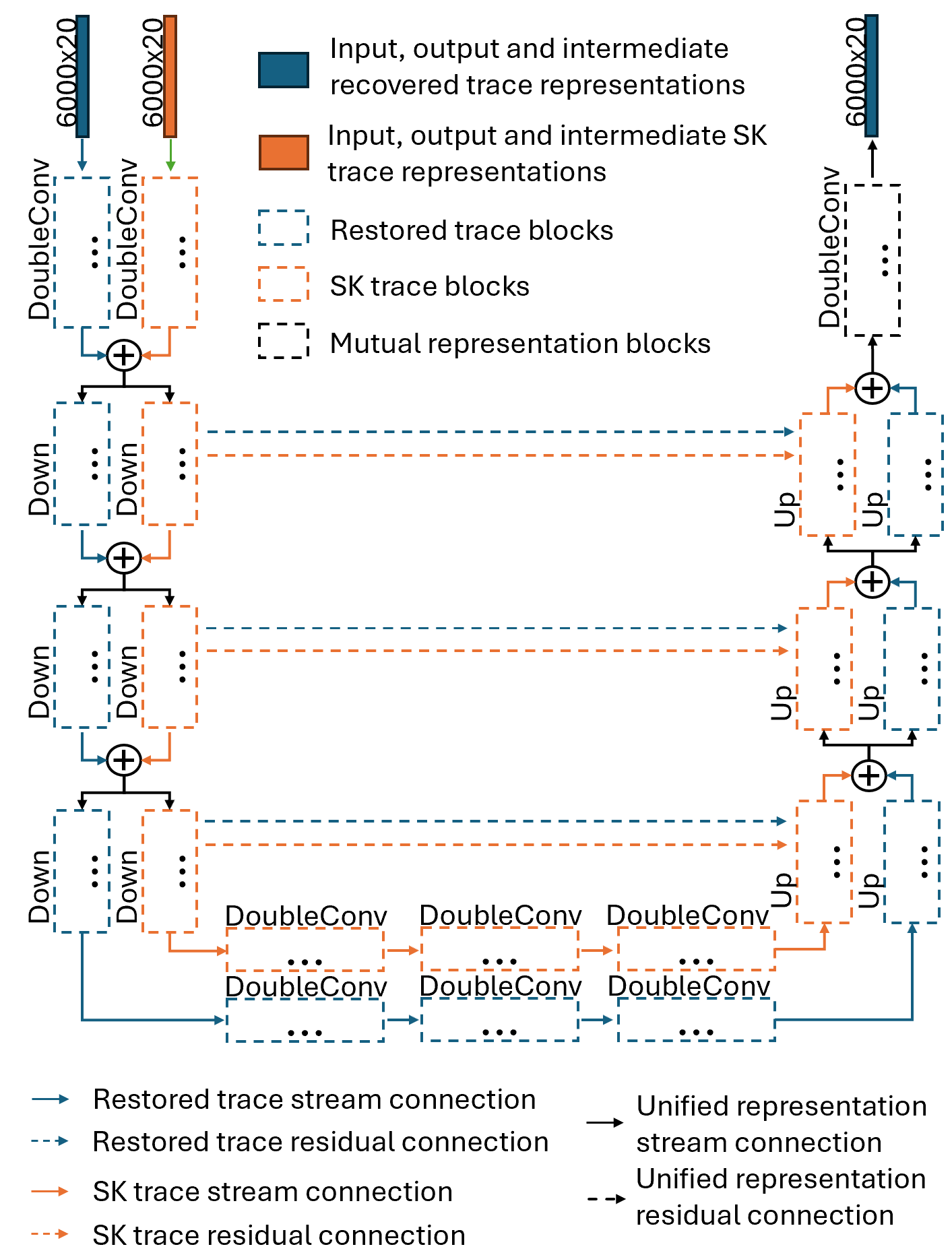}
    \caption{U-net architecture for trace generation conditioned on an SK trace. Each computation stream (blue for the denoised trace and orange for the SK trace) learns separate features, which are added in between blocks such that each block learns its features from the combined representation of the previous blocks. The bottleneck blocks do not combine representations to learn quality high level features for each stream separately. The blue computation stream on its own is identical to the one introduced in Figure~\ref{fig:unet_base}.}
    \label{fig:unet_sk}
\end{figure}

\begin{figure}[t]
    \centering
    \includegraphics[width=1.1\linewidth]{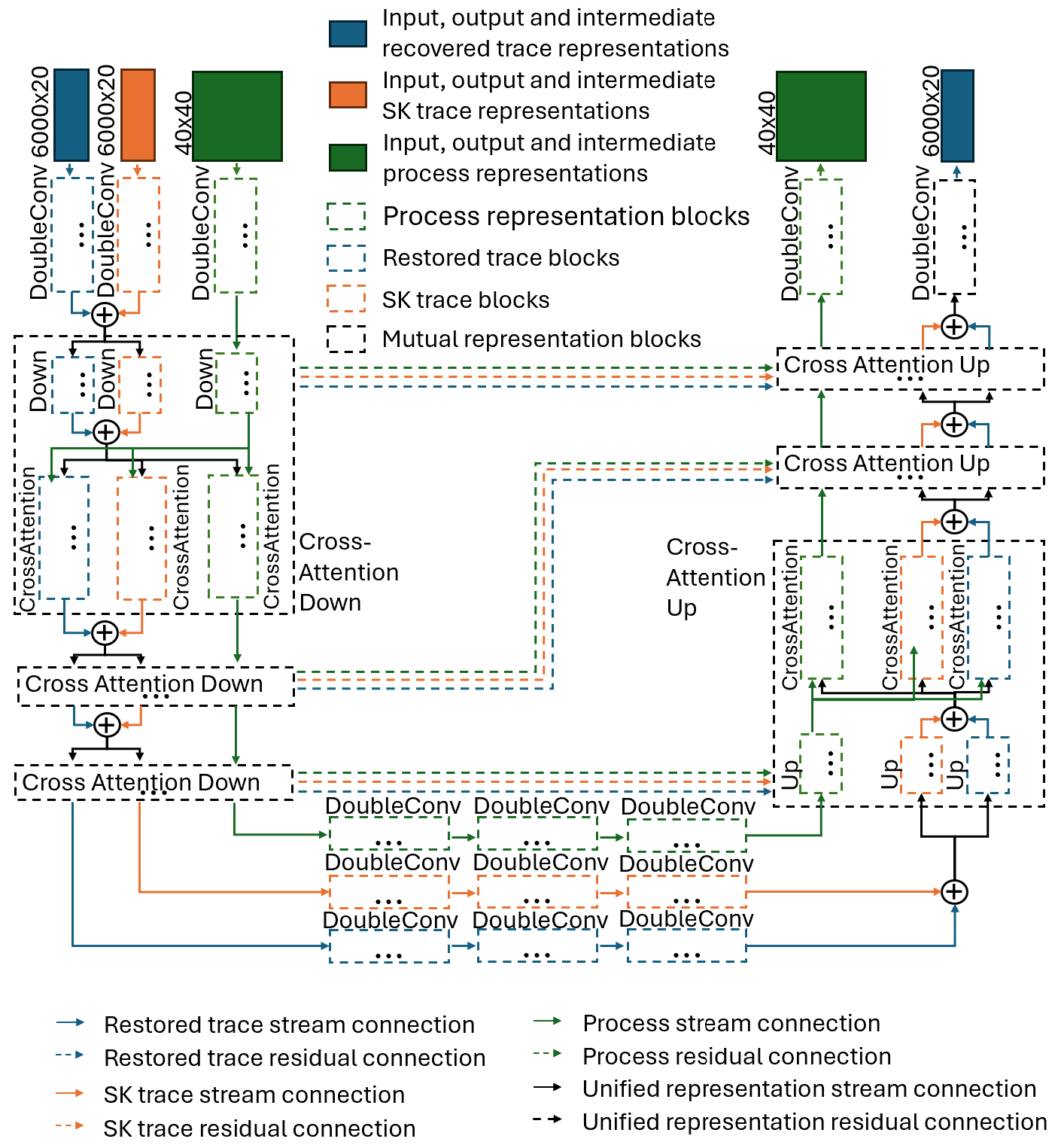}
    \caption{U-net architecture for trace generation conditioned on an SK trace and process model flow matrix. Each computation stream (blue for the denoised trace, orange for the SK trace and green for flow matrix) learns separate features which are added in between blocks such that each block learns its features from the combined representation of the previous blocks. Trace and matrix features are combined by cross attention.}
    \label{fig:unet_sk_pm}
\end{figure}

The architecture of the denoiser neural network is the key component for the success of our method, as it needs to effectively learn representations for traces and the flow matrix and to combine them in a meaningful way. In this work, we expand a U-net (Figure~\ref{fig:unet_base}) to a novel architecture for trace guidance (Figure~\ref{fig:unet_sk}) and combined guidance on a trace and latent process flow matrix (Figure~\ref{fig:unet_sk_pm}).

The base U-net architecture (Figure~\ref{fig:unet_base}) serves as the backbone for our denoiser, it receives a denoised trace from the previous reverse process iteration and outputs the prediction for the DK trace in the current iteration.

The reverse process by itself supports unconditional trace generation once converged, however our goal is to generate a recovered DK trace that is based on a given SK trace.
To achieve this goal we modify the U-net (Figure~\ref{fig:unet_sk}) by adding an input stream for the SK trace (marked in orange), which serves as guidance for the generative process. We duplicate the layers of the base U-net to get two separate computation streams for the denoised trace and the SK trace. Each stream learns its features through the same procedure of convolution and upsampling as described in Section~\ref{sec:Background}.

The separate representations in each block are added together, yielding an intermediate combined representation that is passed forward as input to the next blocks. The final block outputs the predicted DK trace based on the final combined representation. We term this architecture the {\em model-free denoiser} and denote it $H(x_t,y,\varnothing,t;\theta)$ with return values $(\hat{x}_0,\varnothing)$. The model-free denoiser is used to solve Problem~\ref{prob1}. 

To solve Problem~\ref{prob2}, we need to inject process knowledge in the denoiser (Figure~\ref{fig:unet_sk_pm}). The process model provides context about the process' structure, which might not be learned from existing trace features. 
The model holds a matrix of learnable parameters that serves as the model's internal latent version of the process' flow matrix. 

As can be observed in Figure~\ref{fig:unet_sk_pm} (top right), the model constructs its approximation for the ground truth flow matrix in addition to the denoised trace. This way, the denoiser never sees the process explicitly, which allows us to use it in the loss (see Algorithm~\ref{alg:alg_train}), resulting in all model parameters being influenced by the correct identification of the process. For this reason we preferred this approach over giving the actual flow matrix as input, which is also possible with our architecture.


On a technical level, we add an additional computation stream for the latent flow matrix. Combining the process representations with the two trace representations is challenging due to the inherent differences in the representation of the flow matrix. Therefore, simple addition no longer suffices, causing the model to collapse during training. To overcome this challenge, we use cross attention, a deep learning technique that extends self attention. Instead of cross examining a sequence with itself, cross attention examines two distinct sequences. This mechanism is frequently used to bridge different modalities in multi-modal tasks, therefore it is a natural fit when combining trace and process model representations. 

The cross attention mechanism learns to relate parts of the process model and the trace. We use cross attention after each convolution block (Figure~\ref{fig:unet_sk_pm}, top-left), so each representation is influenced by both the combined representation of the denoised trace and the SK trace, and the latent matrix.

The architecture, as presented in Figure~\ref{fig:unet_sk_pm}, is termed the {\em model-aware denoiser}, denoted as $H(x_t,y,\tilde{F},t;\theta)$ with its return values  $(\hat{x}_0,\hat{F})$.

\subsection{Algorithms}
\label{sec:algs}
DDTR involves two distinct phases, executed sequentially, namely {\em training} and {\em inferencing}. 

Algorithm~\ref{alg:alg_train} describes a training phase, during which model parameters are learned using training data.
Lines 4-8 implement the forward process: a random time point along the forward process is randomly selected in line 4, random Gaussian noise is generated in line 5, lines 6-7 randomly drop guidance signals (to improve robustness), with $S$ being the SK trace and $F$ the flow matrix, and in line 8 a perturbed trace is created according to the forward process of Eq.~\ref{eq:forward_proc}. In line 9, the model predicts the original DK trace and flow matrix given the perturbed trace, SK trace, and latent flow matrix. Finally, in line 10 model parameters, $\theta$, are updated according to a gradient descent algorithm. $\theta$ is the set of all model parameters, including weights and biases among all network layers. Each parameter learns a specific feature in the high-dimensional encoding space of the network. During training the entire training dataset is evaluated multiple times according to the number of epochs. Dropping the guidance signals, {\em i.e.}, the SK trace and latent flow matrix, allows more robust feature learning and avoids
memorization of specific examples and overfitting.

\begin{algorithm}[t]
\footnotesize
	\SetKwInOut{Input}{input} \SetKwInOut{Output}{output} \SetKwInOut{Initialization}{initialization}
	\SetAlgoLined
	\Input{dataset $\mathcal{D}$, process model petri net $N$}
	\Output{trained denoiser $H(x_t,y,\tilde{F},t;\theta)$}
	\Initialization{set hyperparameter values: noise schedule $\{\alpha_t\}_{t=1}^T$, number of training epochs $E$, dropout probabilities $p_{no-SK}$, $p_{no-F}$} 
	\BlankLine
	obtain flow matrix $F$ from $N$ \label{algline:adj}\\
	\For {\textbf{each} $\text{epoch} \in \{1,\dots,E\}$\label{algline:train_start}} { 
		\For{\textbf{each} $(\mathbf{D},\mathbf{S})\in\mathcal{D}$\label{algline:epoch_start}} 
		{
			Draw $t\sim \text{Uniform}(\{1,\dots,T\})$\\
                Draw $\varepsilon\sim\mathcal{N}(\vec{0},\mathbf{I})$\\
                $\mathbf{S}\leftarrow\varnothing~\text{w.p}~p_{no-SK}$,~$F\leftarrow\varnothing~\text{w.p}~p_{no-F}$\\
                $x_0\leftarrow\mathbf{D}, y\leftarrow\mathbf{S}$\\
                $x_t\leftarrow\sqrt{\bar{\alpha}_t}x_0+\sqrt{1-\bar{\alpha}_t}\varepsilon$\\
                $(\hat{x}_0, \hat{F})\leftarrow H(x_t,y,\tilde{F},t;\theta)$\\
                Update $\theta$ with a gradient optimization strategy (such as SGD, ADAM) on $\nabla_\theta[\gamma\text{CE}(x_0,\hat{x}_0)+(1-\gamma)\text{CE}(F,\hat{F})]$
		} \label{algline:trained}
        }
        return trained denoiser $H(x_t,y,\tilde{F},t;\theta)$
	\caption{DDTR Training (forward process)}
	\label{alg:alg_train}
\end{algorithm}

Denoiser parameters are updated by a gradient optimization strategy on a loss function that compares the ground truth DK trace and the denoiser's prediction for it. Specifically, we aim to minimize the weighted sum of cross entropy between the recovered trace and the ground truth trace and the cross entropy between the predicted flow matrix and the ground truth matrix. This in turn results in a loss that represents a trade-off between being faithful to the log and the process model and the training phase aims to balance the two.



Notably, DDTR operates in the space of activity log-probabilities rather than directly over categorical DK traces. This design choice enables both DK and SK traces to be represented within a unified continuous space, facilitating effective guidance during the denoising process. Since SK traces naturally encode uncertainty as probability distributions, converting them to log-probabilities (via log transform) aligns seamlessly with this representation. Similarly, DK traces are encoded as one-hot vectors and embedded in the same space. The denoiser outputs logits, which are passed through a softmax transformation to yield probabilities, and the final trace prediction is obtained via argmax selection. This approach is supported by recent findings in diffusion literature~\cite{diffusionce}, which demonstrate its effectiveness in discrete generative tasks.


\begin{algorithm}[t]
\footnotesize

	\SetKwInOut{Input}{input} \SetKwInOut{Output}{output} \SetKwInOut{Initialization}{initialization}
	\SetAlgoLined
	\Input{SK trace $\mathbf{S}$, process model petri net $N$, trained denoiser $H(x_t,y,\tilde{F},t;\theta)$}
	\Output{recovered trace $\mathbf{\hat{D}}$} 
	\BlankLine
        Draw $\hat{x}_T\sim\mathcal{N}(\vec{0},\mathbf{I})$\\
	\For {\textbf{each} $t \in \{T,\dots,1\}$\label{algline:gen_start}} {
            $y\leftarrow\mathbf{S}$\\
            \If {$t>1$} {
                Draw $z\sim\mathcal{N}(\vec{0},\mathbf{I})$\\
            }
            \Else {
                $z\leftarrow0$\\
            }
            $(\hat{x}_0, \hat{F})\leftarrow H(\hat{x}_t,y,\tilde{F},t;\theta)$\\
            $\hat{x}_{t-1}\leftarrow\frac{\sqrt{\bar{\alpha}_{t-1}}}{1-\bar{\alpha}_t}
         \hat{x}_0+
      \frac{\sqrt{\alpha_t}(1-\bar{\alpha}_{t-1})}{1-\bar{\alpha}_t}
         \hat{x}_t+
      \frac{1-\bar{\alpha}_{t-1}}{1-\bar{\alpha}_t}(1-\alpha_t)z $
        }
        $\mathbf{\hat{D}}\leftarrow\text{argmax}(\text{softmax}(\hat{x}_0))$\\
        return recovered trace $\mathbf{\hat{D}}$\\
	\caption{DDTR Inference (reverse process)}
	\label{alg:alg_infer}
\end{algorithm}

The training phase yields a trained denoiser that maximizes the accuracy of DK trace predictions across the training data while also accurately predicting the flow matrix. The trained denoiser is used in the inference phase (Algorithm~\ref{alg:alg_infer}) to effectively recover previously unseen traces and generally any trace that comes from the same process as the training data. 

The inference phase implements the reverse process (Eq.~\ref{eq:back_proc}). It starts with randomly sampled Gaussian noise (line 1) and iterates backward over all denoising steps. In each denoising step (lines 10 and 11) the denoiser takes as input the result of the previous denoising step, as well as the SK trace and latent flow matrix, and output its prediction for the DK trace for the current step. This prediction is then mixed with the previous denoising step and random Gaussian noise and the process repeats until the last denoising step, in which no Gaussian noise is added. The reverse process results in a gradually refined prediction for the DK trace and yields in the end the final recovered trace in log probability form (see Figure~\ref{fig:denoising_example} for example). To recover the trace from probabilities, softmax and argmax functions are applied per activity.

As a final note, algorithm~\ref{alg:alg_train} and~\ref{alg:alg_infer} solve Problem~\ref{prob2}. To solve Problem~\ref{prob1} we modify the algorithms by using the model-free denosier $H(x_t,y,\varnothing,t;\theta)$ and set $F\equiv\varnothing$ and $\gamma=1$.

\section{Empirical Evaluation}
\label{sec:eval}
We evaluate DDTR on both real-world and synthetic datasets. The real-world datasets consist of food preparation datasets: 50-Salads, Breakfast and GTEA, which have been used to benchmark machine learning video action segmentation models.
In these datasets, SK traces are acquired by applying ASFormer~\cite{asformer}, a state-of-the-art video action segmentation model, followed by softmax transformation to the log-probability outputs for each frame to get activity probabilities. DK traces are acquired from the ground truth labels provided in each dataset. Synthetic datasets consist of BPI 2012 and BPI 2019 datasets, DK traces are taken as is and SK traces are generated synthetically using Eq.~\ref{eq:det_to_stoch} with a constant noise level of $0.6$ and uniform concentration parameters of $0.05$. We also generate synthetic variants of the real-world datasets, with noise levels ranging in $[0.53,0.62]$ to evaluate DDTR's robustness. 

In our implementation of DDTR we use the architecture in Figure~\ref{fig:unet_sk_pm}. We use the Adam optimizer with a learning rate of $5\times10^{-6}$ over $5000$ epochs, $500$ diffusion time steps and dropout probabilities of $0.1$ for the SK trace and flow matrix. For each dataset we train the model on $75\%$ of the data and test DDTR and the baselines on the remaining $25\%$. The flow matrix $F$ is taken from a Petri net generated by applying inductive miner on the train set. Models were trained on a server using an RTX Quadro 6000 and CentOS 9 operating system. The code for DDTR is publicly available.\footnote{\url{https://github.com/maxim-mat/DDTR}}

The baselines against which we compare are Argmax, which involves taking the argmax per activity in the SK trace as is, and SKTR~\cite{sktr}. We use the accuracy, macro precision and macro recall as our performance measures.

\subsection{Overall Performance}
Table \ref{tab:overall_results} summarizes the performance of DDTR compared to SKTR and Argmax. We observe that DDTR achieves the highest performance across all datasets, SKTR achieves the second highest performance, followed by argmax. DDTR achieves an improvement in accuracy of 5\%-25\% on the real-world datasets. SKTR performance is closer to DDTR on synthetic datasets, possibly due to the more complex noise distribution in the real-world datasets. 

\begin{table}[htpb]
    \centering
        \caption{Performance metrics on test portion across tested datasets and methods. Bold outlines the highest performing method per metric and dataset. 
        }
    \begin{tabular}{ccccc}
         \hline
         \textbf{Dataset} & \textbf{Method} & \textbf{Accuracy} & \textbf{Precision} & \textbf{Recall} \\
         \hline
         \multirow{3}{*}{50-Salads} & Argmax & 0.774 & 0.744 & 0.731 \\
          & SKTR & 0.89 & 0.83 & 0.81 \\
          & DDTR & \textbf{0.937} & \textbf{0.907} & \textbf{0.9} \\
         \hline
         GTEA & Argmax & 0.751 & 0.751 & 0.755 \\
          & SKTR & 0.79 & 0.819 & 0.782 \\
          & DDTR & \textbf{0.988} & \textbf{0.974} & \textbf{0.969} \\
         \hline
         Breakfast & Argmax & 0.737 & 0.604 & 0.61 \\
          & SKTR & 0.81 & 0.724 & 0.744 \\
          & DDTR & \textbf{0.931} & \textbf{0.876} & \textbf{0.864} \\
          \hline
         BPI 2012 & Argmax & 0.778 & 0.639 & 0.651 \\
          & SKTR & 0.935 & 0.855 & 0.858 \\
          & DDTR & \textbf{0.997} & \textbf{0.994} & \textbf{0.994} \\
          \hline
         BPI 2019 & Argmax & 0.784 & 0.685 & 0.69 \\
          & SKTR & 0.866 & 0.804 & 0.818 \\
          & DDTR & \textbf{0.982} & \textbf{0.973} & \textbf{0.976} \\
    \end{tabular}

    \label{tab:overall_results}
\end{table}

\subsection{Robustness and difference between problems}
To test the robustness of the proposed approach, we vary the level of uncertainty in the SK traces using Eq.~\ref{eq:det_to_stoch} and measure the average accuracy at each noise level. 

\begin{figure}[htpb]
    \centering
    \includegraphics[width=1.2\linewidth]{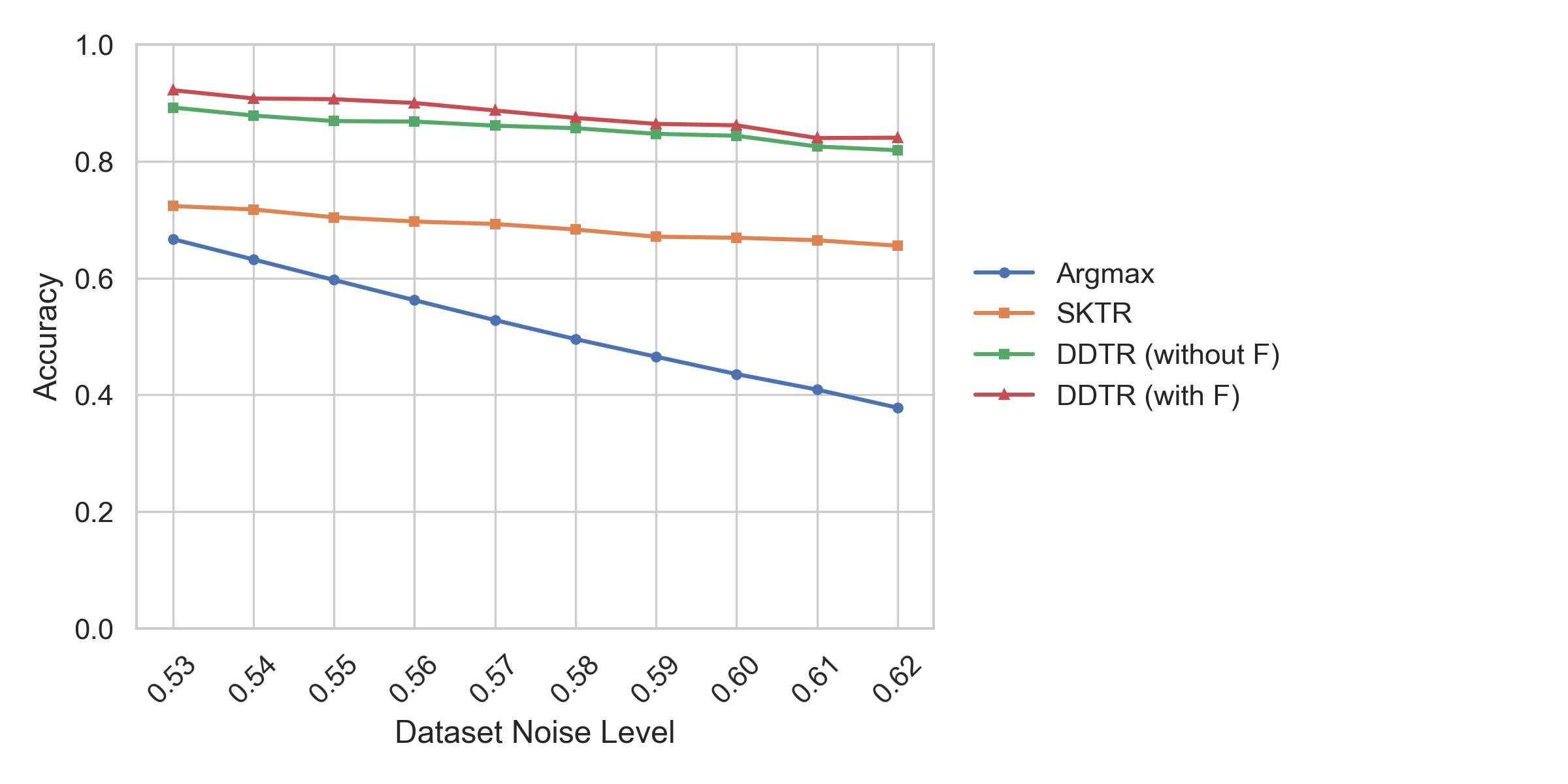}
    \caption{average accuracies of recovery methods across real-world datasets with added synthetic noise. As SK uncertainty increases,  argmax accuracy sharply decreases while DDTR accuracy remains stable. 
    }
    \label{fig:synth_acc}
\end{figure}

Figure~\ref{fig:synth_acc} provides the average accuracy over the three real-world datasets. The accuracy of the argmax approach decreases sharply with the increase in noise levels. SKTR accuracy in this scenario ranges between 0.77 and 0.68. DDTR shows the best performance and also demonstrates a graceful reduction of accuracy as noise levels increase. Model-aware DDTR accuracy stays between 0.96 and 0.83 while model-free DDTR accuracy is between 0.95 and 0.81. We observe that model-free DDTR, which solves the more difficult Problem~\ref{prob1}, achieves accuracy almost on par with the model-aware DDTR, which is exposed to the process model. Synthetic benchmarking was done without any additional training or fine-tuning of either DDTR model on synthetic noise variants.

\section{Related Work}
\label{sec:rel}
Our work focuses on accurately recovering DK traces from SK traces using a machine learning approach,
which can be viewed as a type of a log repair approach. Hence, we divide the related work into two main bodies of works, SK trace recovery methods (Section~\ref{sec:relatedSK}), and works on log repair (Section~\ref{sec:relatedLogML}).

\subsection{Trace Recovery}\label{sec:relatedSK}
Bogdanov {\em et al}.~\cite{sktr} tackled the problem of SK trace recovery (in essence solving Problem~\ref{prob2}). In their work, dubbed SKTR, the authors use a multigraph search approach where edges are weighted by a cost function proportional to the SK probabilities. The resulting shortest path is a sequence of activities that maximizes the joint probability in the multigraph. While SKTR achieve significant improvement in accuracy over the argmax baseline, our approach surpasses its performance in all test scenarios. Additionally, the runtime of SKTR increases with trace length, as the size of the constructed multigraph grows along with the trace. On the other hand, DDTR leverages a neural network which runs in constant time per batch, along with a predefined constant number of denoising steps, which ensures consistent runtime.
\subsection{Log Repair}\label{sec:relatedLogML}
Machine learning methods were effectively used for various log repair tasks. Wu {\em et al}.~\cite{logrepair_transformer} uses a masked transformer model to substitute missing and anomalous activities in event logs. The model is trained on a simulated dataset that is generated by injecting missing values and anomalies to an existing high-quality event log. Lu {\em et al.}~\cite{missing_activities} uses an LSTM-based prediction model to predict missing prefix and suffix sequences of activity labels in otherwise high-quality event logs. Dentamaro {\em et al}.~\cite{suffix_lstm} trade-off a multi-speed transformer and a bidirectional LSTM for the tasks of next activity prediction and elapsed time prediction. Fang and Su~\cite{clustering_repair} use spectral clustering to repair missing and anomalous data on both an attribute and case level. Unlike DDTR, these approaches operate on deterministic logs with missing or erroneous data, making direct comparison difficult. However, an interesting direction for future work would be to adapt DDTR for repairing errors in stochastically known logs, where uncertainty is explicit.

Beyond machine learning-based approaches, Rogge-Solti {\em et al.}~\cite{DBLP:conf/otm/Rogge-SoltiMAW13} propose a method for repairing event logs using stochastic process models, specifically stochastic Petri nets. While DDTR similarly incorporates model structure into its recovery mechanism, it operates in a fundamentally different space—using deep generative models rather than alignment-based techniques. Combining such model-based alignment with data-driven generative methods may open up promising directions for future research in stochastic log repair.

\section{Conclusion}
\label{sec:conclusion}

This work introduced DDTR, a novel method for recovering deterministically known traces from stochastically known process logs using guided diffusion denoising models. By reframing trace recovery as an inverse problem and leveraging the flexibility of Diffusion Denoising Probabilistic Models (DDPMs), we present a deep learning approach capable of reconstructing traces with high accuracy and robustness. The method generalizes DDPM guidance to include both probabilistic trace data and process model structure, grounding the recovery in both local uncertainty and global constraints.

DDTR outperforms prior methods across real-world and synthetic datasets, improving accuracy by up to 25\%. It remains robust under increasing noise and handles traces of arbitrary length without reliance on alignment-based search. Incorporating a latent representation of the process model enhances recovery in high uncertainty scenarios, especially when event probabilities become unreliable.

Future work may extend DDTR to incorporate timestamps, resources, or natural language annotations. Another direction involves supporting concurrent or hierarchical process structures beyond sequential traces ({\em e.g.}, process trees and event structures). Finally, integrating DDTR into full process mining pipelines would enable end-to-end handling of uncertain event data in practical applications.

\bibliographystyle{IEEEtran}
\bibliography{IEEEabrv,references}

\end{document}